\documentclass[pdflatex,sn-mathphys-num]{sn-jnl}

\usepackage{amsmath,amssymb,amsfonts}
\usepackage{graphicx}
\usepackage{booktabs}
\usepackage{array}
\usepackage{url}
\usepackage{microtype}
\usepackage{xcolor}
\usepackage{tikz}
\usetikzlibrary{arrows.meta,positioning,fit,calc}
\usepackage{pgfplots}
\usepgfplotslibrary{groupplots}
\pgfplotsset{compat=1.18}

\makeatletter
\renewcommand\normalsize{%
  \@setfontsize\normalsize{10bp}{12bp}%
  \abovedisplayskip 6\p@ \@plus2\p@ \@minus2\p@
  \abovedisplayshortskip \z@ \@plus2\p@
  \belowdisplayshortskip 3\p@ \@plus2\p@ \@minus2\p@
  \belowdisplayskip \abovedisplayskip
  \let\@listi\@listI}
\renewcommand\section{\@startsection{section}{1}{\z@}%
  {-8pt \@plus-2pt \@minus-1pt}{4pt}{\sectionfont}}
\renewcommand\subsection{\@startsection{subsection}{2}{\z@}%
  {-7pt \@plus-2pt \@minus-1pt}{3pt}{\subsectionfont}}
\renewcommand\subsubsection{\@startsection{subsubsection}{3}{\z@}%
  {-6pt \@plus-2pt \@minus-1pt}{2pt}{\subsubsectionfont}}
\makeatother
\normalsize
\definecolor{metricaccuracy}{HTML}{0077BB}
\definecolor{metricfscore}{HTML}{EE7733}
\definecolor{methodagriarc}{HTML}{0077BB}
\definecolor{methodhific}{HTML}{009988}
\definecolor{methodbyol}{HTML}{33BBEE}
\definecolor{methoddenoising}{HTML}{EE7733}
\definecolor{methodmae}{HTML}{EE3377}
\definecolor{methodsimclr}{HTML}{CC3311}
\definecolor{methodcbrl}{HTML}{6A3D9A}

\begin{document}

\title[Contrastive Bootstrap Representation Learning]{Contrastive Bootstrap Representation Learning for Label-Efficient Agricultural Image Classification}

\author[1]{\fnm{Hazem} \sur{Ahmed}}
\author[1]{\fnm{Ali} \sur{Hamdi}}
\author[1]{\fnm{Ammar} \sur{Mohammed}}

\affil[1]{\orgdiv{Department of Computer Science},
  \orgname{Faculty of Graduate Studies for Statistical Research, Cairo University},
  \orgaddress{\city{Cairo}, \country{Egypt}}}

\abstract{
Expert annotation remains a major bottleneck in agricultural image classification, where collecting images is often easier than obtaining reliable labels from domain specialists. Self-supervised learning can reduce this dependence by pretraining transferable representations on unlabelled imagery. Contrastive and bootstrap methods provide two established but distinct training signals, motivating an empirical study of a formulation that applies both signals within one representation path under severe downstream label scarcity.

We present \textbf{Contrastive Bootstrap Representation Learning (CBRL)}, which jointly optimises SimCLR-style~\cite{chen2020simclr} NT-Xent contrastive learning and bidirectional BYOL-style~\cite{grill2020byol} exponential-moving-average target prediction within a shared online projection space. Using a ViT-L/16 backbone~\cite{dosovitskiy2021vit} and 10\% of the downstream labels, we report single-seed point estimates on four primary agricultural test benchmarks plus an exploratory 33-image New Plant Diseases evaluation. CBRL records the highest macro-F1 point estimate on the four primary benchmarks. On DeepWeeds~\cite{olsen2018deepweeds}, its 79.77\% test macro-F1 is 32.29 points above the random-initialisation, label-only baseline and 15.78 points above the next-highest comparison point estimate; on PlantVillage~\cite{hughes2015plantvillage}, its 98.14\% macro-F1 is 10.43 and 0.43 points higher, respectively. On PlantDoc, however, all methods fail to reach practically useful performance. These results identify the evaluated joint formulation as a promising candidate for label-limited transfer, but they do not establish objective synergy, statistical superiority, or robustness across stochastic runs.
}


\maketitle

\section{Introduction}

Agricultural image classification is an important application of computer vision in which the annotation bottleneck is particularly pronounced. Images can be collected at scale from field surveys, mobile diagnostic tools, phenotyping systems, and online agricultural platforms, yet obtaining reliable labels often requires domain expertise. This challenge is especially evident in plant disease recognition, where annotations may vary across crops, disease categories, growth stages, and acquisition conditions. PlantVillage~\cite{hughes2015plantvillage} was released as an open image collection to support mobile plant disease diagnostics and contains more than 50,000 curated crop-leaf images. Deep neural classifiers have demonstrated that such datasets can support automated disease recognition~\cite{mohanty2016plantdisease}. However, supervised learning assumes access to sufficient labelled data and therefore does not address a broader challenge faced by agricultural vision systems: how effectively can large amounts of unlabelled agricultural imagery be exploited when downstream annotations are scarce?

Self-supervised learning (SSL) provides a natural approach to this problem by learning transferable visual representations without requiring task-specific labels during pretraining. In general computer vision, methods such as SimCLR~\cite{chen2020simclr}, BYOL~\cite{grill2020byol}, Barlow Twins~\cite{zbontar2021barlow}, DINO~\cite{caron2021dino}, DINOv2~\cite{oquab2023dinov2}, and MAE~\cite{he2022mae} have demonstrated that carefully designed self-supervised objectives can produce representations that transfer effectively to downstream tasks. These methods, however, optimise substantially different learning signals. Contrastive approaches such as SimCLR~\cite{chen2020simclr} encourage inter-instance discrimination by attracting representations of augmented views of the same image while separating different images in a batch. Bootstrap approaches such as BYOL~\cite{grill2020byol} instead promote cross-view consistency by predicting representations generated by a slowly updated target network, without explicitly relying on negative samples. Other approaches rely on redundancy reduction, self-distillation, or masked reconstruction. Although these objectives have been extensively studied on large-scale natural-image benchmarks, their joint use for label-limited agricultural representation learning remains insufficiently characterised.

This question is particularly relevant because agricultural imagery can differ substantially from the natural-image benchmarks on which many SSL methods have been developed. Agricultural classification may depend on subtle texture variations, localised lesions, colour changes, plant morphology, and spatial patterns that distinguish visually similar categories. Consequently, representations that perform well for generic object recognition may not necessarily capture the features most useful for agricultural transfer. Vision Transformers (ViTs)~\cite{dosovitskiy2021vit} provide a suitable backbone for investigating this problem because their patch-token representation supports contrastive, bootstrap, redundancy-reduction, masked-reconstruction, and dense-feature objectives within a common architecture family. However, the effectiveness of jointly exploiting contrastive discrimination and bootstrap-based cross-view consistency within a controlled ViT architecture and under a limited-label downstream protocol remains insufficiently characterised.

To address this gap, we empirically evaluate a joint batch-contrastive and target-bootstrap formulation for label-limited agricultural image classification. We propose \textbf{Contrastive Bootstrap Representation Learning (CBRL)}, which uses shared online representations for SimCLR-style~\cite{chen2020simclr} discrimination against other images in the batch and bidirectional prediction of a slowly updated BYOL-style~\cite{grill2020byol} target representation. This design places both training signals in one representation path; whether they interact synergistically is not tested by the present study. We evaluate the complete formulation using a ViT-Large backbone on PlantVillage~\cite{hughes2015plantvillage}, DeepWeeds~\cite{olsen2018deepweeds}, PlantDoc~\cite{singh2019plantdoc}, and LUCAS Cover Photos~\cite{dandrimont2020lucas}, with New Plant Diseases~\cite{bhattarai2018newplant} retained as an exploratory evaluation because its prepared test split contains only 33 images. All downstream runs use 10\% of the available training labels under a common fine-tuning protocol.

The main contributions of this work are as follows:
\begin{itemize}
\setlength{\itemsep}{0pt}
\setlength{\parskip}{0pt}
\setlength{\parsep}{0pt}
\item We specify a reproducible joint formulation that applies NT-Xent batch contrast and bidirectional EMA-target prediction to a shared online ViT encoder and projection space.
\item We provide a broad single-seed, point-estimate comparison of ten initialisation strategies under a common 10\%-label downstream protocol on four primary agricultural benchmarks.
\item We characterize both favourable and adverse outcomes: large observed margins on DeepWeeds, near-saturation on PlantVillage, failure-level absolute performance on PlantDoc, and a separately qualified exploratory result on the incomplete New Plant Diseases test split.
\end{itemize}

\section{Related Work}

\subsection{Label-Efficient Agricultural Recognition}

Agricultural image recognition is particularly affected by limited labelled data because reliable annotation often requires domain expertise, while image acquisition itself can be comparatively inexpensive. Plant disease recognition is a prominent example: PlantVillage~\cite{hughes2015plantvillage} introduced a large open collection of healthy and diseased crop-leaf images for mobile disease diagnostics, and Mohanty et al.\ demonstrated the feasibility of deep supervised classification on PlantVillage~\cite{mohanty2016plantdisease}. However, agricultural recognition extends beyond controlled leaf imagery. The datasets considered in this work span controlled or augmented crop images in PlantVillage~\cite{hughes2015plantvillage} and New Plant Diseases~\cite{bhattarai2018newplant}, rangeland weed imagery in DeepWeeds~\cite{olsen2018deepweeds}, web-sourced disease images in PlantDoc~\cite{singh2019plantdoc}, and survey photographs in LUCAS~\cite{dandrimont2020lucas,yordanov2023lucas}. Differences in acquisition conditions, backgrounds, taxonomies, and class distributions make cross-dataset evaluation important for assessing whether learned representations transfer beyond a single curated benchmark.

Several approaches have sought to reduce the dependence of agricultural recognition systems on large labelled datasets. Nuthalapati and Tunga~\cite{nuthalapati2021multidomain} investigate multi-domain few-shot recognition across pests, plants, and diseases and introduce a real-world agricultural few-shot dataset. Within self-supervised learning, Wang et al.~\cite{wang2024plantssl} combine masked-autoencoder pretraining with attention mechanisms for leaf-disease classification. More closely related to multi-objective representation learning, Huan et al.~\cite{huan2025unified} combine BYOL, masked-image modelling, and contrastive learning using a ResNet-101 backbone on laboratory and field plant-disease imagery. Collectively, these studies demonstrate the relevance of label-efficient and self-supervised learning to agricultural vision. However, differences in task scope, backbone architecture, pretraining objective, and evaluation protocol make it difficult to isolate how alternative self-supervised objectives behave under the same downstream conditions. This work therefore evaluates them within a controlled Vision Transformer architecture~\cite{dosovitskiy2021vit} and limited-label transfer setting.

\subsection{Contrastive and Non-Contrastive Vision Pretraining}

Self-supervised vision methods differ primarily in how they construct invariance and prevent representational collapse. Contrastive methods learn representations by attracting augmented views of the same image while separating representations of different images. SimCLR~\cite{chen2020simclr} implements this principle through the NT-Xent objective, in-batch negatives, and a carefully designed augmentation pipeline. Momentum Contrast (MoCo)~\cite{he2020moco} instead maintains a queue of negative keys together with a moving-averaged key encoder, thereby decoupling the contrastive dictionary from the current mini-batch. Importantly, MoCo~\cite{he2020moco} demonstrates that an exponential-moving-average encoder can serve a contrastive objective and is therefore not exclusive to negative-free bootstrap learning.

Non-contrastive methods remove explicit instance negatives but employ different mechanisms to avoid collapse and preserve informative representations. BYOL~\cite{grill2020byol} trains an online predictor against stop-gradient targets produced by an exponential-moving-average target network, whereas SimSiam~\cite{chen2021simsiam} demonstrates that stop-gradient optimisation can support Siamese representation learning without either negative samples or a momentum encoder. Barlow Twins~\cite{zbontar2021barlow} instead reduces redundancy by encouraging the cross-correlation matrix between paired views to approach the identity matrix. DINO~\cite{caron2021dino} and DINOv2~\cite{oquab2023dinov2} employ teacher--student self-distillation to learn transferable ViT representations, while MAE~\cite{he2022mae} learns representations through reconstruction of masked image patches with an asymmetric encoder/decoder architecture. These methods therefore encode different inductive biases through instance discrimination, cross-view prediction, redundancy reduction, self-distillation, and masked reconstruction. Comparing them under a common downstream protocol helps separate differences in learning objective from differences introduced by evaluation settings.

\subsection{Joint Contrastive and Teacher--Student Objectives}

Prior work has combined contrastive learning with momentum encoders or teacher--student mechanisms in several forms. MoCo~\cite{he2020moco} uses a momentum encoder to produce consistent keys while retaining a contrastive dictionary-lookup objective. ReSSL~\cite{zheng2021ressl} also employs a momentum teacher, but learns by matching sharpened distributions of inter-instance similarities rather than by combining NT-Xent with a direct BYOL-style prediction objective. In agricultural vision, Huan et al.~\cite{huan2025unified} integrate BYOL, masked reconstruction, and contrastive learning as distinct components within a unified framework. Consequently, the contribution of CBRL is not the general combination of contrastive learning with a momentum or teacher--student branch, but the specific manner in which the two learning signals are coupled within a shared representation path.

CBRL applies two explicit objectives to the same online representation space. The online projections participate in SimCLR-style NT-Xent discrimination, while an online predictor aligns them bidirectionally with opposite-view projections produced by a stop-gradient EMA target network. Unlike MoCo~\cite{he2020moco}, CBRL does not maintain a negative queue; unlike ReSSL~\cite{zheng2021ressl}, it does not match relational target distributions; and unlike the framework of Huan et al.~\cite{huan2025unified}, it does not introduce a masked-reconstruction branch. The contribution evaluated here is this specific complete formulation and a controlled downstream benchmark, rather than a claim that its terms interact synergistically. The parent-method comparisons contextualise its point estimates but, without strict loss-toggle ablations, do not identify the causal contribution of either term.

ArcFace-SSL and HiFiC-SSL are additionally included as project-specific comparison methods in the benchmark. ArcFace-SSL applies ArcFace-style~\cite{deng2019arcface} angular-margin instance discrimination as a way of using metric-learning loss functions in a self-supervised manner, whereas HiFiC-SSL adapts the rate--distortion perceptual adversarial formulation used in high-fidelity generative compression~\cite{mentzer2020hific} to compress the input into a latent space that carries useful information about the input. These methods are included as comparative baselines rather than components of the proposed CBRL framework.

\section{Methods}

\subsection{Study Design}

We conduct an empirical comparison of downstream fine-tuning performance under alternative random and self-supervised initialisation strategies. Each SSL method produces a pretrained ViT~\cite{dosovitskiy2021vit} encoder, which is subsequently fine-tuned on four primary benchmarks---PlantVillage~\cite{hughes2015plantvillage}, DeepWeeds~\cite{olsen2018deepweeds}, PlantDoc~\cite{singh2019plantdoc}, and LUCAS Cover Photos~\cite{dandrimont2020lucas}---and on the exploratory New Plant Diseases evaluation~\cite{bhattarai2018newplant}. Classification is the only downstream task considered.

The reference condition is a random-initialisation, label-only baseline. It follows the same downstream protocol but receives no representation pretraining. Its purpose is to measure the observed benefit of SSL initialisation over training this large backbone from scratch with limited labels; it is not presented as a strongest-possible supervised transfer baseline, because ImageNet-supervised initialisation and supervised hyperparameter optimisation were not evaluated. Each method has one completed run per dataset with 10\% of the available downstream training labels. All reported comparisons are therefore descriptive, single-seed point estimates without uncertainty intervals or significance tests.

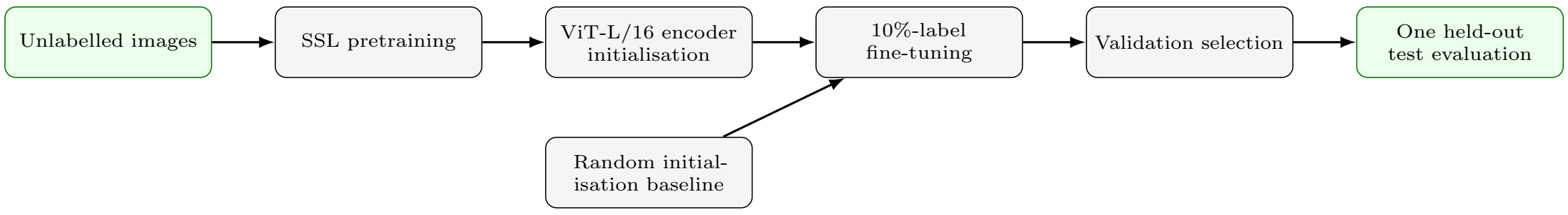
\begin{figure}[!t]
\centering
\resizebox{0.90\textwidth}{!}{%
\begin{tikzpicture}[
    font=\footnotesize,
    node distance=0.8cm,
    block/.style={draw, rounded corners, align=center, minimum height=0.9cm, text width=2.4cm, fill=gray!8},
    data/.style={draw=green!45!black, rounded corners, align=center, minimum height=0.9cm, text width=2.4cm, fill=green!8},
    arrow/.style={-{Latex[length=2mm]}, thick}
]
\node[data] (unlab) {Unlabelled images};
\node[block, right=of unlab] (ssl) {SSL pretraining};
\node[block, right=of ssl] (encoder) {ViT-L/16 encoder initialisation};
\node[block, right=of encoder] (tune) {10\%-label fine-tuning};
\node[block, right=of tune] (select) {Validation selection};
\node[data, right=of select] (test) {One held-out test evaluation};
\node[block, below=0.75cm of encoder] (random) {Random initialisation baseline};
\draw[arrow] (unlab) -- (ssl);
\draw[arrow] (ssl) -- (encoder);
\draw[arrow] (encoder) -- (tune);
\draw[arrow] (random) -- (tune);
\draw[arrow] (tune) -- (select);
\draw[arrow] (select) -- (test);
\end{tikzpicture}%
}
\caption{Common evaluation pipeline. SSL methods provide a ViT-L/16 initialisation; the label-only reference bypasses pretraining and begins from random weights. Each selected checkpoint is evaluated once, so the reported values are point estimates.}
\label{fig:experimental-pipeline}
\end{figure}

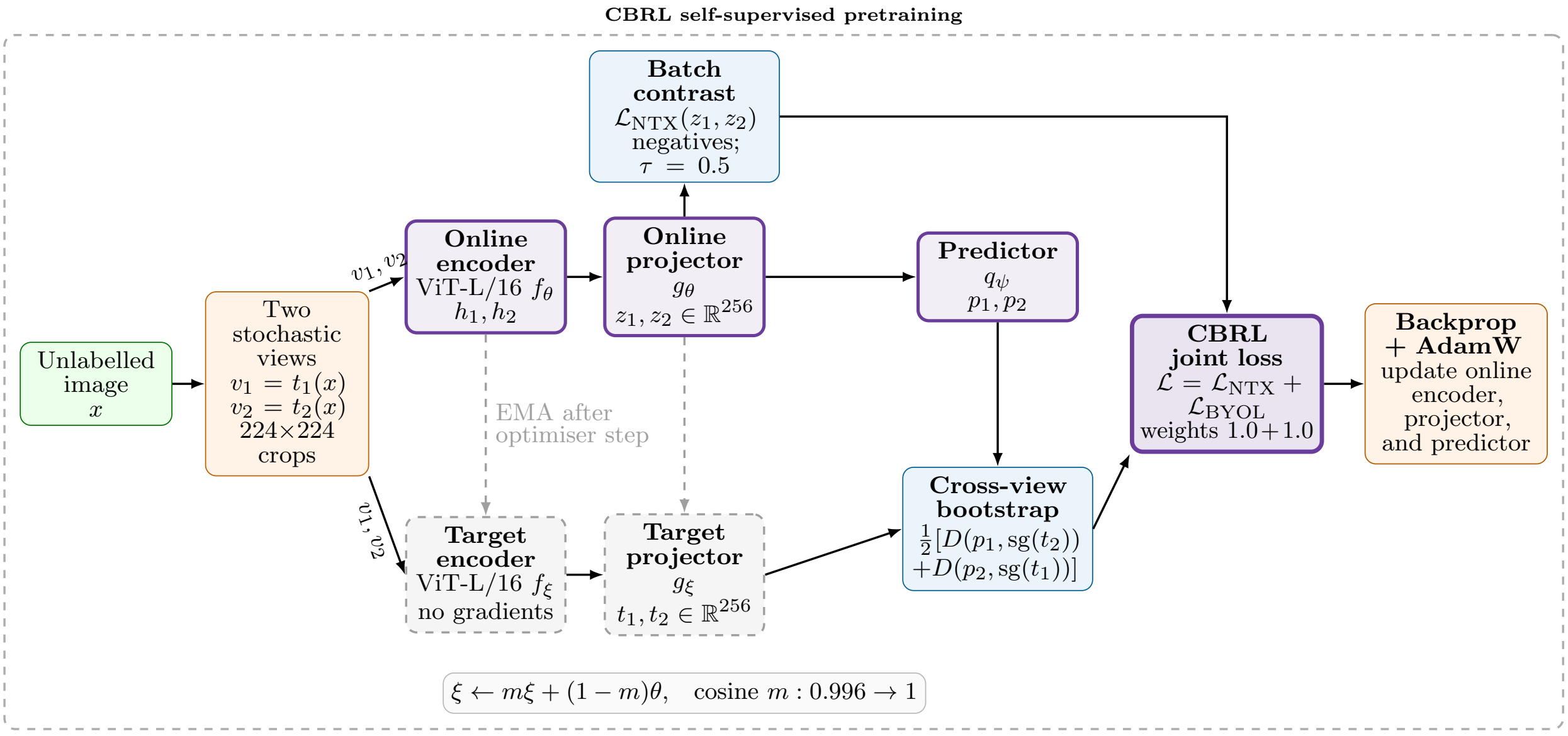
\begin{figure}[!t]
\centering
\resizebox{\textwidth}{!}{%
\begin{tikzpicture}[
    font=\scriptsize,
    data/.style={draw=green!45!black, rounded corners, align=center, minimum height=0.9cm, text width=1.75cm, fill=green!8},
    transform/.style={draw=orange!75!black, rounded corners, align=center, minimum height=0.9cm, text width=1.9cm, fill=orange!10},
    online/.style={draw=methodcbrl, very thick, rounded corners, align=center, minimum height=0.95cm, text width=1.85cm, fill=methodcbrl!9},
    target/.style={draw=gray!75, dashed, thick, rounded corners, align=center, minimum height=0.95cm, text width=1.85cm, fill=gray!8},
    objective/.style={draw=metricaccuracy!80!black, rounded corners, align=center, minimum height=0.9cm, text width=2.25cm, fill=metricaccuracy!8},
    joint/.style={draw=methodcbrl, ultra thick, rounded corners, align=center, minimum height=1.0cm, text width=2.25cm, fill=methodcbrl!14},
    train/.style={draw=orange!70!black, rounded corners, align=center, minimum height=0.9cm, text width=2.15cm, fill=orange!10},
    result/.style={draw=green!45!black, rounded corners, align=center, minimum height=0.9cm, text width=1.95cm, fill=green!8},
    note/.style={draw=gray!55, rounded corners, align=center, inner sep=3pt, fill=gray!4},
    arrow/.style={-{Latex[length=1.8mm]}, thick},
    ema/.style={-{Latex[length=1.8mm]}, dashed, thick, color=gray!75},
    stage/.style={draw=gray!65, rounded corners, dashed, thick, inner sep=0.2cm}
]
\node[data] (unlab) at (0,3.5) {Unlabelled image\\$x$};
\node[transform] (aug) at (2.5,3.5) {Two stochastic views\\$v_1=t_1(x)$\\$v_2=t_2(x)$\\$224{\times}224$ crops};

\node[online] (oenc) at (5.1,4.9) {\textbf{Online encoder}\\ViT-L/16 $f_\theta$\\$h_1,h_2$};
\node[online] (oproj) at (7.7,4.9) {\textbf{Online projector}\\$g_\theta$\\$z_1,z_2\in\mathbb{R}^{256}$};
\node[online] (pred) at (11.8,4.9) {\textbf{Predictor}\\$q_\psi$\\$p_1,p_2$};

\node[target] (tenc) at (5.1,1.0) {\textbf{Target encoder}\\ViT-L/16 $f_\xi$\\no gradients};
\node[target] (tproj) at (7.7,1.0) {\textbf{Target projector}\\$g_\xi$\\$t_1,t_2\in\mathbb{R}^{256}$};

\node[objective] (ntx) at (7.7,7.0) {\textbf{Batch contrast}\\$\mathcal{L}_{\mathrm{NTX}}(z_1,z_2)$\\negatives; $\tau=0.5$};
\node[objective] (boot) at (11.8,1.6) {\textbf{Cross-view bootstrap}\\$\frac{1}{2}[D(p_1,\mathrm{sg}(t_2))$\\$+D(p_2,\mathrm{sg}(t_1))]$};
\node[joint] (jointloss) at (14.8,3.5) {\textbf{CBRL joint loss}\\$\mathcal{L}=\mathcal{L}_{\mathrm{NTX}}+\mathcal{L}_{\mathrm{BYOL}}$\\weights $1.0+1.0$};
\node[train] (step) at (17.8,3.5) {\textbf{Backprop + AdamW}\\update online encoder,\\projector, and predictor};

\draw[arrow] (unlab) -- (aug);
\draw[arrow] (aug.north east) -- node[above,sloped] {$v_1,v_2$} (oenc.west);
\draw[arrow] (aug.south east) -- node[below,sloped] {$v_1,v_2$} (tenc.west);
\draw[arrow] (oenc) -- (oproj);
\draw[arrow] (oproj) -- (pred);
\draw[arrow] (tenc) -- (tproj);
\draw[arrow] (oproj.north) -- (ntx.south);
\draw[arrow] (pred.south) -- (boot.north);
\draw[arrow] (tproj.east) -- (boot.west);
\draw[arrow] (ntx.east) -- ++(0.65,0) -| (jointloss.north);
\draw[arrow] (boot.east) -- (jointloss.south west);
\draw[arrow] (jointloss) -- (step);

\draw[ema] (oenc.south) -- node[right,align=left] {\scriptsize EMA after\\[-1pt]\scriptsize optimiser step} (tenc.north);
\draw[ema] (oproj.south) -- (tproj.north);
\node[note] (emaf) at (7.7,-0.55) {$\xi\leftarrow m\xi+(1-m)\theta$,\quad cosine $m:0.996\rightarrow1$};

\node[stage, fit=(unlab)(aug)(oenc)(oproj)(pred)(tenc)(tproj)(ntx)(boot)(jointloss)(step)(emaf),
      label={[font=\footnotesize\bfseries]above:CBRL self-supervised pretraining}] (pretrainbox) {};
\end{tikzpicture}%
}
\caption{CBRL pretraining mechanism. NT-Xent uses the online projections; bidirectional bootstrap prediction uses opposite-view stop-gradient targets. Their fixed 1.0/1.0 weighted sum updates the online path, while the target path follows by EMA. Figure~\ref{fig:experimental-pipeline} shows the separate downstream evaluation.}
\label{fig:architecture}
\end{figure}

Figure~\ref{fig:experimental-pipeline} provides the high-level pretraining, transfer, fine-tuning, model-selection, and evaluation workflow used across the benchmark. Figure~\ref{fig:architecture} then details the internal CBRL pretraining mechanism without replacing this common experimental pipeline.

Figure~\ref{fig:architecture} summarises the gradient and EMA update paths during CBRL pretraining and identifies the components retained for downstream transfer.

\subsection{Datasets and Split Protocol}

The downstream benchmark comprises four primary agricultural image-classification datasets and one exploratory dataset. PlantVillage~\cite{hughes2015plantvillage} is a curated collection developed for mobile plant-health diagnostics and contains more than 50,000 healthy and diseased crop-leaf images in the original release. The evaluated split contains 38 classes and 3,803 labelled training examples. New Plant Diseases~\cite{bhattarai2018newplant} is an augmented plant-leaf dataset containing approximately 87,000 RGB images across 38 classes, with a source-provided 80/20 train-validation organisation. Although its evaluated training split contains 7,030 labelled images, the prepared test split contains only 33 images and omits many output classes. We therefore classify this dataset as exploratory and exclude it from the primary cross-dataset claims.

DeepWeeds~\cite{olsen2018deepweeds} is a public multiclass weed-recognition dataset collected in Australian rangelands. Its source paper reports 17,509 labelled images covering eight nationally significant weed species collected across eight locations. The evaluated split contains nine classes, including the negative class, and 1,225 labelled training examples. PlantDoc~\cite{singh2019plantdoc} is a real-world plant-disease dataset constructed from internet-sourced imagery; the source paper reports 2,598 data points covering 13 plant species and up to 17 disease classes. In the evaluated split, crop disease combinations define 28 classes, with 235 labelled training examples. LUCAS Cover Photos~\cite{dandrimont2020lucas} are derived from the EU Land Use/Cover Area frame Survey, a geolocated in-situ land-cover and land-use survey containing harmonised observations and millions of associated photographs. The evaluated split uses a 59-class cover taxonomy and contains 13,669 labelled training examples, rather than the 12-crop subset examined by Yordanov et al.~\cite{yordanov2023lucas}.

For each dataset, a deterministic subset containing 10\% of the original training labels was constructed. Validation and held-out test partitions were held fixed across methods for all five datasets. Table~\ref{tab:split-summary} summarises the resulting class counts and numbers of labelled training examples.

\begin{table}[t]
\caption{Classification split summary for the 10\%-label evaluation.}
\label{tab:split-summary}
\centering
\footnotesize
\begin{tabular*}{\textwidth}{@{\extracolsep{\fill}}lrr@{}}
\toprule
Dataset & Classes & Labelled training examples \\
\midrule
PlantVillage & 38 & 3,803 \\
New Plant Diseases$^\dagger$ & 38 & 7,030 \\
DeepWeeds & 9 & 1,225 \\
PlantDoc & 28 & 235 \\
LUCAS Cover Photos & 59 & 13,669 \\
\botrule
\end{tabular*}
\par\raggedright\scriptsize $^\dagger$Exploratory evaluation; the prepared test split contains only 33 images and incomplete class coverage.
\end{table}

The datasets represent distinct agricultural recognition settings. PlantVillage~\cite{hughes2015plantvillage}, New Plant Diseases~\cite{bhattarai2018newplant}, and PlantDoc~\cite{singh2019plantdoc} focus on crop-disease recognition; DeepWeeds~\cite{olsen2018deepweeds} focuses on weed-species recognition in rangeland imagery; and LUCAS~\cite{dandrimont2020lucas} focuses on land-cover or crop-cover classification. We therefore report both accuracy and macro-F1 because class frequencies vary substantially across the prepared datasets.

\subsection{Model and Fine-Tuning Protocol}

All reported runs use a ViT-Large~\cite{dosovitskiy2021vit} encoder with an input resolution of $224 \times 224$. For downstream classification, the encoder representation is passed through dropout followed by a fully connected linear classification head.

Fine-tuning uses AdamW with separate learning rates for the encoder and classifier head. The encoder learning rate is $1.0 \times 10^{-5}$, the head learning rate is $1.0 \times 10^{-4}$, and the weight decay is 0.05. The training and evaluation batch sizes are 64 and 128, respectively. Each model is trained for 50 epochs with five warm-up epochs, cosine learning-rate scheduling, and automatic mixed precision on compatible hardware. Model selection is based on the highest validation macro-F1 obtained during training.

Accuracy and macro-F1 are used as downstream evaluation metrics, with macro-F1 serving as the primary model-selection criterion because it weights all classes equally. For each dataset, the checkpoint achieving the highest validation macro-F1 is selected and evaluated once on the held-out test set.

\subsection{Unified Backbone and Downstream Objective}

All SSL methods are implemented as alternative pretraining objectives for the same encoder family. Let $x$ denote an agricultural image and let $f_\theta(x) \in \mathbb{R}^{d}$ denote the global representation produced by the encoder. During SSL pretraining, $\theta$ denotes the trainable backbone parameters. Method-specific projectors, predictors, prototypes, decoders, discriminators, and teacher networks are used only to define their respective pretraining objectives and are discarded before downstream evaluation.

For downstream classification, a supervised head $g_\phi$ is attached to the pretrained encoder:
\begin{equation}
    \hat{y} = g_\phi(f_{\theta^\star}(x)),
\end{equation}
where $\theta^\star$ denotes either SSL-pretrained parameters or the randomly initialised parameters used by the label-only reference condition. Fine-tuning minimises cross-entropy over the selected labelled training partition:
\begin{equation}
    \mathcal{L}_{\mathrm{cls}}(\theta,\phi) =
    -\frac{1}{N_l}\sum_{i=1}^{N_l}\log
    \frac{\exp(g_\phi(f_\theta(x_i))_{y_i})}
    {\sum_{c=1}^{C}\exp(g_\phi(f_\theta(x_i))_{c})}.
    \label{eq:cls}
\end{equation}
Both the backbone and classifier head are trainable during downstream fine-tuning. Their learning rates are set to $1.0 \times 10^{-5}$ and $1.0 \times 10^{-4}$, respectively, allowing the classifier to adapt more rapidly to the downstream label space while the transferred encoder receives smaller updates.

\subsection{Pretraining Methods Compared}

The downstream comparison includes the proposed CBRL method and nine alternative initialisation strategies:
\begin{enumerate}
\setlength{\itemsep}{0pt}
\setlength{\parskip}{0pt}
\setlength{\parsep}{0pt}
    \item Random-initialisation, label-only ViT baseline~\cite{dosovitskiy2021vit}: no SSL pretraining; downstream training uses only the selected labelled subset.
    \item SimCLR~\cite{chen2020simclr}: two-view contrastive pretraining using the NT-Xent objective.
    \item Barlow Twins~\cite{zbontar2021barlow}: two-view redundancy reduction based on cross-correlation alignment.
    \item BYOL~\cite{grill2020byol}: online-to-target prediction using an exponential-moving-average target network.
    \item DINOv2~\cite{oquab2023dinov2}: self-distillation with masked patch prediction and KoLeo regularisation.
    \item MAE~\cite{he2022mae}: masked-patch reconstruction.
    \item Denoising autoencoding~\cite{vincent2008denoising}: reconstruction of clean images from corrupted inputs.
    \item ArcFace-SSL: instance-level ArcFace-style angular-margin SSL with view consistency and entropy balancing~\cite{deng2019arcface}.
    \item HiFiC-SSL: compression-inspired SSL combining rate, distortion, perceptual, and adversarial objectives~\cite{mentzer2020hific}.
    \item \textbf{CBRL (ours)}: joint NT-Xent contrastive learning and bidirectional EMA-target prediction within a shared projection space~\cite{chen2020simclr,grill2020byol}.
\end{enumerate}

CBRL uses 256-dimensional online and target projections and 4096-dimensional hidden layers in both the projector and predictor. The NT-Xent temperature is $\tau=0.5$, and the target-network momentum begins at 0.996 and increases towards 1 according to a cosine schedule. The contrastive and bootstrap loss weights were fixed at 1.0 as a simple, non-tuned design choice. Equal coefficients do not imply equal gradient influence or optimal weighting.

The saved configurations and TensorBoard logs confirm that, within each dataset-specific comparison, CBRL, SimCLR, and BYOL used a ViT-L/16 backbone, the same unlabelled images, 10,000 optimiser steps, batch size 256, and the common downstream protocol. BYOL is the closest parent comparison because it also matches CBRL's online/target architecture, projector and predictor dimensions, EMA schedule, optimiser, and augmentations, while omitting the contrastive term. SimCLR retains its method-specific projector, LARS optimiser, and augmentation settings. Consequently, these comparisons control important resources but are not strict loss-toggle ablations.

\subsection{SSL Objective Summary}

The standard comparison objectives are summarised here; their complete formulations follow the cited original works. SimCLR~\cite{chen2020simclr} uses NT-Xent with in-batch negatives, Barlow Twins~\cite{zbontar2021barlow} aligns cross-correlations while reducing redundancy, and BYOL~\cite{grill2020byol} predicts stop-gradient targets from an EMA network. MAE~\cite{he2022mae} reconstructs masked patches, denoising autoencoding~\cite{vincent2008denoising} reconstructs clean inputs from corrupted images, and DINOv2~\cite{oquab2023dinov2} combines teacher--student self-distillation, masked patch prediction, and feature regularisation. We retain the equations below only for CBRL and the two project-specific comparison objectives, for which implementation detail is needed to distinguish the evaluated formulations.

\subsection{Proposed Contrastive Bootstrap Representation Learning}

CBRL combines contrastive discrimination and bootstrap target prediction within a shared online representation path. Given two views $v_1$ and $v_2$, the online encoder $f_\theta$ and projector $g_\theta$ produce
\begin{equation}
    z_i = g_\theta(f_\theta(v_i)), \qquad i\in\{1,2\},
\end{equation}
and the online predictor $q_\psi$ produces
\begin{equation}
    p_i = q_\psi(z_i).
\end{equation}
A target encoder and projector with parameters $\xi$ produce target representations $z_i^\xi$. No gradients are propagated through the target branch.

The CBRL objective is
\begin{equation}
    \begin{aligned}
    \mathcal{L}_{\mathrm{CBRL}} =
    &\lambda_{\mathrm{con}}\mathcal{L}_{\mathrm{NTXent}}(z_1,z_2)\\
    &+\frac{\lambda_{\mathrm{boot}}}{2}\left[
    D(p_1,\mathrm{sg}(z_2^\xi))+
    D(p_2,\mathrm{sg}(z_1^\xi))\right],
    \end{aligned}
    \label{eq}
\end{equation}
where
\begin{equation}
    D(a,b)=2-2\cos(a,b),
\end{equation}
$\lambda_{\mathrm{con}}=\lambda_{\mathrm{boot}}=1.0$, and the NT-Xent temperature is $\tau=0.5$. The first term applies instance-level contrastive discrimination to the online projections, while the second applies bidirectional cross-view prediction between the online and target branches.

After each optimiser step, the target parameters are updated by an exponential moving average:
\begin{equation}
    \xi \leftarrow m_t\xi+(1-m_t)\theta,
    \label{eq:ema}
\end{equation}
where $m_t$ follows a cosine schedule from an initial value of 0.996 towards 1. Gradients update only the online encoder, projector, and predictor. The target encoder and projector are updated exclusively through Eq.~\ref{eq:ema}.

The complete objective places explicit cross-instance separation and target prediction in one training formulation. Its two coefficients were fixed at 1.0 without a weighting search. CBRL records higher downstream point estimates than both parent methods in the reported runs. However, BYOL is the only closely matched architectural comparison, while SimCLR retains method-native pretraining choices; neither comparison supplies repeated runs. The evidence therefore evaluates the complete formulation but does not prove synergy or isolate the causal contribution of either loss.

\subsection{Project Comparison Objectives}

\subsubsection{ArcFace-SSL}
ArcFace-SSL treats each unlabelled training image as an instance identity during pretraining. For image $i$, two augmented views produce embeddings $h_{i,1}$ and $h_{i,2}$. The ArcFace classifier~\cite{deng2019arcface} maintains normalised prototype weights $w_c$ and computes cosine logits. For the positive instance class $y_i=i$, the angular-margin logit is
\begin{equation}
    s \cos(\theta_{y_i}+m),
    \quad
    \theta_{y_i}=\arccos
    \left(
    \frac{w_{y_i}^{\top}h_i}{\|w_{y_i}\|_2\|h_i\|_2}
    \right),
    \label{eq:arcface}
\end{equation}
where $s=30.0$ and $m=0.3$ denote the scale and angular margin, respectively. The two-view instance-classification term is
\begin{equation}
    \mathcal{L}_{\mathrm{arc}} =
    \frac{1}{2}\left[
    \mathrm{CE}(A(h_{i,1}), y_i) +
    \mathrm{CE}(A(h_{i,2}), y_i)
    \right],
\end{equation}
where $A(\cdot)$ denotes the ArcFace logit function. The complete ArcFace-SSL objective additionally includes view-consistency and entropy-balancing terms:
\begin{equation}
    \begin{aligned}
    \mathcal{L}_{\mathrm{ArcFace\mbox{-}SSL}} =
    &\mathcal{L}_{\mathrm{arc}}
    + \lambda_{\mathrm{con}}
    \left(1-\cos(h_{i,1},h_{i,2})\right) \\
    &+ \lambda_{\mathrm{ent}}
    \sum_c \bar{p}_c \log(\bar{p}_c).
    \end{aligned}
    \label{eq:agriarc}
\end{equation}
The corresponding weights are $\lambda_{\mathrm{con}}=1.0$ and $\lambda_{\mathrm{ent}}=0.1$. Because each pseudo-label corresponds to an image identity rather than a crop or disease class, no semantic class labels are used during pretraining.

\subsubsection{HiFiC-SSL}
HiFiC-SSL adapts high-fidelity learned compression to self-supervised representation learning. The ViT backbone produces patch-level features that are projected to a latent tensor $y$. A hyper-analysis transform produces $z$; quantisation is approximated by uniform noise during training; and a hyper-synthesis transform predicts scale parameters for a Gaussian likelihood model. A synthesis decoder reconstructs the image from the quantised latent $\hat{y}$.

The generator objective is
\begin{equation}
    \mathcal{L}_{\mathrm{HiFiC\mbox{-}SSL}} =
    \lambda_d D(x,\hat{x}) +
    \lambda_r R(\hat{y},\hat{z}) +
    \lambda_p P(x,\hat{x}) +
    \lambda_a A_G(\hat{x}),
    \label{eq:hific}
\end{equation}
where $D$ denotes image distortion, $R$ is a bitrate proxy computed from latent likelihoods, $P$ is a frozen DINOv2 class-token perceptual distance, and $A_G$ is the generator adversarial loss. The implementation uses mean-squared-error distortion with
$\lambda_d=1.0$, $\lambda_r=0.01$, $\lambda_p=0.1$, and $\lambda_a=0.01$.

The patch discriminator is trained using logistic real/fake losses:
\begin{equation}
    \mathcal{L}_{D} =
    \mathbb{E}_{x}[\mathrm{softplus}(-D_\omega(x))] +
    \mathbb{E}_{\hat{x}}[\mathrm{softplus}(D_\omega(\hat{x}))].
    \label{eq:hific_disc}
\end{equation}
Unlike ArcFace-SSL, HiFiC-SSL does not use an instance-discrimination objective. Its pretrained representation is evaluated solely through the common downstream fine-tuning protocol.
\section{Benchmark Results}

\subsection{Benchmark Coverage Across Evaluation Splits}

The benchmark comprises four primary datasets, one exploratory dataset, and ten initialisation strategies: CBRL, eight SSL comparison methods, and a baseline initialised from random weights. This yields 40 primary and 10 exploratory method--dataset point estimates. Table~\ref{tab:coverage} summarises this coverage.

\begin{table}[t]
\caption{Benchmark coverage for the 10\%-label evaluation.}
\label{tab:coverage}
\centering
\footnotesize
\begin{tabular*}{\textwidth}{@{\extracolsep{\fill}}lrr@{}}
\toprule
Dataset & Evaluations & Test \\
\midrule
PlantVillage & 10 & 10 \\
New Plant Diseases$^\dagger$ & 10 & 10 \\
DeepWeeds & 10 & 10 \\
PlantDoc & 10 & 10 \\
LUCAS & 10 & 10 \\
\midrule
Total & 50 & 50 \\
\botrule
\end{tabular*}
\par\raggedright\scriptsize $^\dagger$Exploratory; not counted as primary evidence because the test split has 33 images and incomplete class coverage.
\end{table}

\subsection{Completed Downstream Classification Results}

Table~\ref{tab:classification-results-grid} reports single-seed held-out test point estimates. Panels (a)--(d) are the primary benchmarks; panel (e), New Plant Diseases~\cite{bhattarai2018newplant}, is exploratory. Accuracy (Acc.) and macro-F1 (F1) are percentages. For each method, the checkpoint with the highest validation macro-F1 is selected and evaluated once on the corresponding fixed test partition; the table therefore conveys neither run-to-run uncertainty nor statistical significance.

\begin{table}[!t]
\caption{Single-seed downstream test point estimates using 10\% of the labels and ViT-Large. Bold values denote the largest observed value within each dataset, not statistical superiority. Panel (e) is exploratory.}
\label{tab:classification-results-grid}
\centering
\scriptsize
\renewcommand{\arraystretch}{0.82}
\begin{minipage}{\textwidth}
\begin{minipage}[t]{0.49\textwidth}
\centering
\textbf{(a) PlantVillage}\par\smallskip
\begin{tabular*}{\linewidth}{@{\extracolsep{\fill}}lrr@{}}
\toprule
Method & Acc. & F1 \\
\midrule
Random-init ViT~\cite{dosovitskiy2021vit} & 90.89 & 87.71 \\
\textbf{CBRL (ours)}~\cite{chen2020simclr,grill2020byol} & \textbf{98.70} & \textbf{98.14} \\
ArcFace-SSL~\cite{deng2019arcface} & 93.41 & 90.86 \\
HiFiC-SSL~\cite{mentzer2020hific} & 93.32 & 90.76 \\
DINOv2~\cite{oquab2023dinov2} & 91.08 & 87.43 \\
BYOL~\cite{grill2020byol} & 94.91 & 92.70 \\
Denoising~\cite{vincent2008denoising} & 94.65 & 92.77 \\
MAE~\cite{he2022mae} & 95.54 & 93.61 \\
SimCLR~\cite{chen2020simclr} & 97.31 & 96.34 \\
Barlow Twins~\cite{zbontar2021barlow} & 98.23 & 97.71 \\
\botrule
\end{tabular*}
\end{minipage}\hfill
\begin{minipage}[t]{0.49\textwidth}
\centering
\textbf{(b) DeepWeeds}\par\smallskip
\begin{tabular*}{\linewidth}{@{\extracolsep{\fill}}lrr@{}}
\toprule
Method & Acc. & F1 \\
\midrule
Random-init ViT~\cite{dosovitskiy2021vit} & 61.71 & 47.49 \\
\textbf{CBRL (ours)}~\cite{chen2020simclr,grill2020byol} & \textbf{85.04} & \textbf{79.77} \\
ArcFace-SSL~\cite{deng2019arcface} & 65.66 & 51.22 \\
HiFiC-SSL~\cite{mentzer2020hific} & 69.59 & 57.72 \\
DINOv2~\cite{oquab2023dinov2} & 62.05 & 42.50 \\
BYOL~\cite{grill2020byol} & 67.00 & 54.05 \\
Denoising~\cite{vincent2008denoising} & 73.20 & 63.99 \\
MAE~\cite{he2022mae} & 65.89 & 50.67 \\
SimCLR~\cite{chen2020simclr} & 68.52 & 57.40 \\
Barlow Twins~\cite{zbontar2021barlow} & 52.00 & 7.60 \\
\botrule
\end{tabular*}
\end{minipage}

\vspace{0.35em}

\begin{minipage}[t]{0.49\textwidth}
\centering
\textbf{(c) PlantDoc}\par\smallskip
\begin{tabular*}{\linewidth}{@{\extracolsep{\fill}}lrr@{}}
\toprule
Method & Acc. & F1 \\
\midrule
Random-init ViT~\cite{dosovitskiy2021vit} & 10.17 & 8.06 \\
\textbf{CBRL (ours)}~\cite{chen2020simclr,grill2020byol} & \textbf{11.86} & \textbf{9.31} \\
ArcFace-SSL~\cite{deng2019arcface} & 9.32 & 6.30 \\
HiFiC-SSL~\cite{mentzer2020hific} & 5.08 & 2.91 \\
DINOv2~\cite{oquab2023dinov2} & 8.90 & 6.26 \\
BYOL~\cite{grill2020byol} & 5.08 & 1.64 \\
Denoising~\cite{vincent2008denoising} & 8.47 & 5.45 \\
MAE~\cite{he2022mae} & 5.93 & 1.10 \\
SimCLR~\cite{chen2020simclr} & 10.17 & 7.73 \\
Barlow Twins~\cite{zbontar2021barlow} & 5.08 & 0.35 \\
\botrule
\end{tabular*}
\end{minipage}\hfill
\begin{minipage}[t]{0.49\textwidth}
\centering
\textbf{(d) LUCAS Cover Photos}\par\smallskip
\begin{tabular*}{\linewidth}{@{\extracolsep{\fill}}lrr@{}}
\toprule
Method & Acc. & F1 \\
\midrule
Random-init ViT~\cite{dosovitskiy2021vit} & 44.03 & 12.78 \\
\textbf{CBRL (ours)}~\cite{chen2020simclr,grill2020byol} & 58.01 & \textbf{27.31} \\
ArcFace-SSL~\cite{deng2019arcface} & 50.96 & 20.22 \\
HiFiC-SSL~\cite{mentzer2020hific} & 48.98 & 14.89 \\
DINOv2~\cite{oquab2023dinov2} & 38.03 & 5.32 \\
BYOL~\cite{grill2020byol} & 55.44 & 21.07 \\
Denoising~\cite{vincent2008denoising} & 49.80 & 17.09 \\
MAE~\cite{he2022mae} & \textbf{58.07} & 20.94 \\
SimCLR~\cite{chen2020simclr} & 52.28 & 21.79 \\
Barlow Twins~\cite{zbontar2021barlow} & 53.03 & 18.45 \\
\botrule
\end{tabular*}
\end{minipage}

\vspace{0.35em}

\hfill
\begin{minipage}[t]{0.49\textwidth}
\centering
\textbf{(e) New Plant Diseases (exploratory)}\par\smallskip
\begin{tabular*}{\linewidth}{@{\extracolsep{\fill}}lrr@{}}
\toprule
Method & Acc. & F1 \\
\midrule
Random-init ViT~\cite{dosovitskiy2021vit} & 90.91 & 19.99 \\
\textbf{CBRL (ours)}~\cite{chen2020simclr,grill2020byol} & \textbf{100.00} & \textbf{21.05} \\
ArcFace-SSL~\cite{deng2019arcface} & 93.94 & 20.23 \\
HiFiC-SSL~\cite{mentzer2020hific} & 93.94 & 20.32 \\
DINOv2~\cite{oquab2023dinov2} & 27.27 & 6.32 \\
BYOL~\cite{grill2020byol} & 96.97 & 20.53 \\
Denoising~\cite{vincent2008denoising} & 96.97 & 20.76 \\
MAE~\cite{he2022mae} & \textbf{100.00} & \textbf{21.05} \\
SimCLR~\cite{chen2020simclr} & 96.97 & 20.76 \\
Barlow Twins~\cite{zbontar2021barlow} & 90.91 & 19.76 \\
\botrule
\end{tabular*}
\end{minipage}
\hfill\null
\end{minipage}
\end{table}

Across the four primary datasets, CBRL records the largest macro-F1 point estimate in each single run. It records the largest accuracy point estimate on three and is 0.06 points below MAE~\cite{he2022mae} on LUCAS Cover Photos~\cite{dandrimont2020lucas}. On the exploratory New Plant Diseases evaluation it ties MAE in both metrics. These are numerical rankings of completed runs, not estimates of expected performance across seeds.

PlantVillage~\cite{hughes2015plantvillage} yields the highest absolute test performance. CBRL records 98.70\% accuracy and 98.14\% macro-F1, 7.81 and 10.43 points above the random-initialisation baseline. Its margins over Barlow Twins~\cite{zbontar2021barlow} are only 0.47 and 0.43 points; without repeated runs, this near-saturation difference should not be treated as a reliable advantage.

DeepWeeds~\cite{olsen2018deepweeds} shows the largest observed margins. CBRL records 85.04\% accuracy and 79.77\% macro-F1, 23.33 and 32.29 points above the random-initialisation baseline and 11.84 and 15.78 points above denoising, the strongest comparison run. Barlow Twins~\cite{zbontar2021barlow} records 52.00\% accuracy but 7.60\% macro-F1. This anomalous gap is reported as observed; without collapse diagnostics or reruns, its cause cannot be assigned to the general method.

PlantDoc~\cite{singh2019plantdoc} is a negative result. CBRL has the largest point estimates (11.86\% accuracy and 9.31\% macro-F1), only 1.69 and 1.25 points above the random-initialisation baseline. Every method remains at failure-level absolute performance, so this benchmark provides no evidence of practical usefulness under the evaluated 10\%-label setting.

On LUCAS Cover Photos~\cite{yordanov2023lucas}, CBRL records 27.31\% macro-F1, 14.53 points above the random-initialisation baseline and 5.52 above SimCLR~\cite{chen2020simclr}, the next-largest point estimate. Its 58.01\% accuracy is 0.06 points below MAE~\cite{he2022mae}. The accuracy--macro-F1 gap describes uneven aggregate and class-balanced performance, but it does not identify which classes benefit.

On the exploratory 33-image New Plant Diseases~\cite{bhattarai2018newplant} split, CBRL and MAE~\cite{he2022mae} both record 100.00\% accuracy and 21.05\% macro-F1. The macro-F1 implementation averages over all 38 output classes, including classes absent from this test split. The result is retained for transparency but excluded from primary evidence and from claims of broad generalisation.

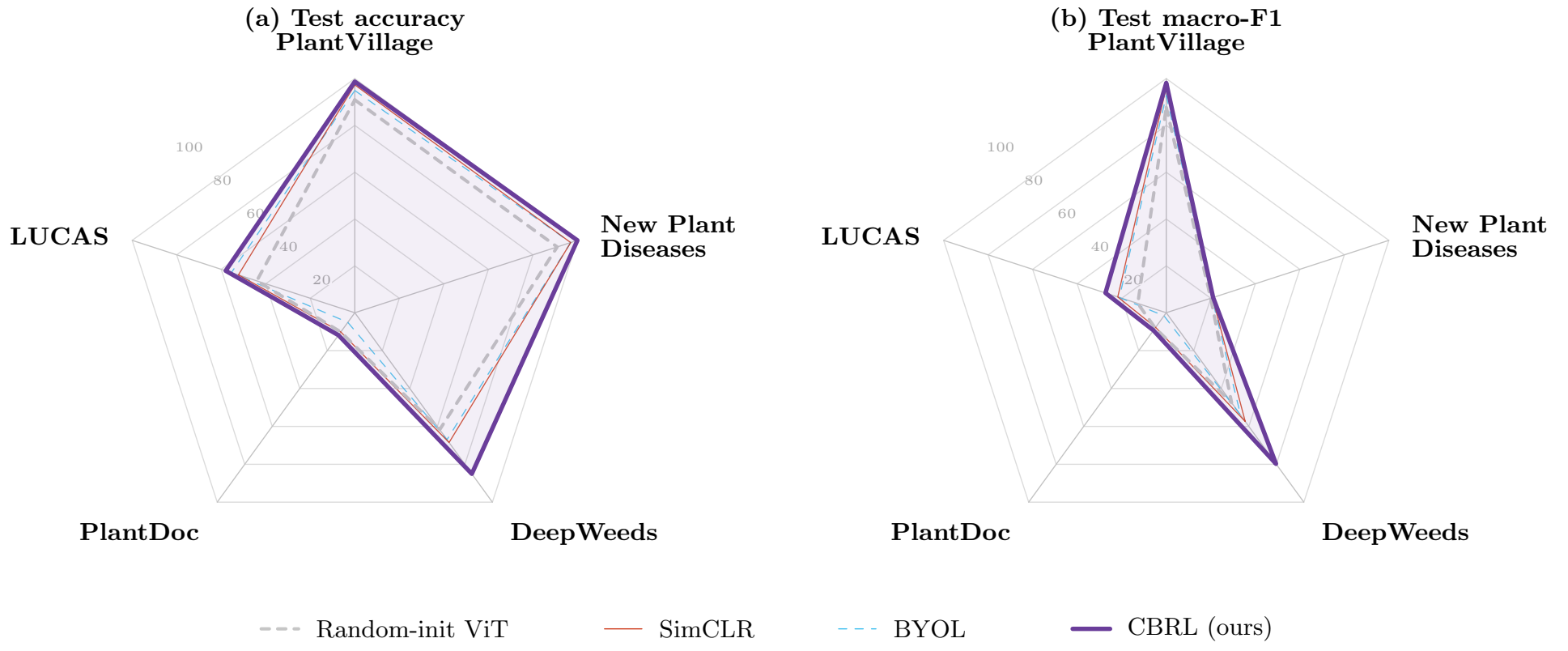
\begin{figure}[!t]
\centering
\resizebox{0.78\textwidth}{!}{%
\begin{tikzpicture}[line join=round,line cap=round]
\def\RadarRadius{3.0}
\newcommand{\radargrid}[1]{%
\begin{scope}[shift={#1}]
\foreach \score in {20,40,60,80,100}{
\draw[gray!28,thin]
(90:{\RadarRadius*\score/100}) --
(18:{\RadarRadius*\score/100}) --
(-54:{\RadarRadius*\score/100}) --
(-126:{\RadarRadius*\score/100}) --
(162:{\RadarRadius*\score/100}) -- cycle;
\node[font=\tiny,text=gray!70,fill=white,inner sep=0.4pt]
at (135:{\RadarRadius*\score/100}) {\score};
}
\foreach \angle in {90,18,-54,-126,162}
\draw[gray!48,thin] (0,0) -- (\angle:\RadarRadius);
\node[font=\scriptsize\bfseries,anchor=south] at (90:{\RadarRadius+0.18}) {PlantVillage};
\node[font=\scriptsize\bfseries,anchor=west,align=left] at (18:{\RadarRadius+0.18}) {New Plant\\Diseases};
\node[font=\scriptsize\bfseries,anchor=north west] at (-54:{\RadarRadius+0.18}) {DeepWeeds};
\node[font=\scriptsize\bfseries,anchor=north east] at (-126:{\RadarRadius+0.18}) {PlantDoc};
\node[font=\scriptsize\bfseries,anchor=east] at (162:{\RadarRadius+0.18}) {LUCAS};
\end{scope}
}
\newcommand{\radarseries}[7]{%
\begin{scope}[shift={#1}]
\draw[#2]
(90:{\RadarRadius*#3/100}) --
(18:{\RadarRadius*#4/100}) --
(-54:{\RadarRadius*#5/100}) --
(-126:{\RadarRadius*#6/100}) --
(162:{\RadarRadius*#7/100}) -- cycle;
\end{scope}
}
\newcommand{\legenditem}[4]{%
\draw[#1] (#2,#3) -- ++(0.48,0);
\node[font=\scriptsize,anchor=west] at ($(#2,#3)+(0.58,0)$) {#4};
}

\coordinate (acc) at (-5.20,0);
\coordinate (fscore) at (5.20,0);
\radargrid{(acc)}
\radargrid{(fscore)}
\node[font=\small\bfseries] at (-5.20,3.75) {(a) Test accuracy};
\node[font=\small\bfseries] at (5.20,3.75) {(b) Test macro-F1};

\radarseries{(acc)}{draw=methodbyol,dashed,thin,opacity=0.82}{94.91}{96.97}{67.00}{5.08}{55.44}
\radarseries{(acc)}{draw=methodsimclr,thin,opacity=0.82}{97.31}{96.97}{68.52}{10.17}{52.28}
\radarseries{(acc)}{draw=gray!45,very thick,dashed}{90.89}{90.91}{61.71}{10.17}{44.03}
\radarseries{(acc)}{draw=methodcbrl,ultra thick,fill=methodcbrl,fill opacity=0.08}{98.70}{100.00}{85.04}{11.86}{58.01}

\radarseries{(fscore)}{draw=methodbyol,dashed,thin,opacity=0.82}{92.70}{20.53}{54.05}{1.64}{21.07}
\radarseries{(fscore)}{draw=methodsimclr,thin,opacity=0.82}{96.34}{20.76}{57.40}{7.73}{21.79}
\radarseries{(fscore)}{draw=gray!45,very thick,dashed}{87.71}{19.99}{47.49}{8.06}{12.78}
\radarseries{(fscore)}{draw=methodcbrl,ultra thick,fill=methodcbrl,fill opacity=0.08}{98.14}{21.05}{79.77}{9.31}{27.31}

\legenditem{draw=gray!45,very thick,dashed}{-6.4}{-4.05}{Random-init ViT}
\legenditem{draw=methodsimclr,thin}{-2.0}{-4.05}{SimCLR}
\legenditem{draw=methodbyol,dashed,thin}{1.0}{-4.05}{BYOL}
\legenditem{draw=methodcbrl,ultra thick}{4.0}{-4.05}{CBRL (ours)}
\end{tikzpicture}
}
\caption{Single-seed test point estimates for CBRL, its SimCLR and BYOL parent-method comparisons, and the random-initialisation reference. The complete ten-method comparison is reported in Table~\ref{tab:classification-results-grid}. New Plant Diseases is exploratory.}
\label{fig:all-testsets-10percent}

\end{figure}

Figure~\ref{fig:all-testsets-10percent} visualises CBRL, its two parent-method comparisons, and the random-initialisation reference. Table~\ref{tab:classification-results-grid} retains the complete ten-method results. The exploratory New Plant Diseases spoke is retained but separated in interpretation.

\subsection{Observed Label-Limited Transfer}

Relative to the random-initialisation baseline under the same 10\%-label protocol, CBRL's macro-F1 point estimate is higher by 10.43 points on PlantVillage~\cite{hughes2015plantvillage}, 32.29 on DeepWeeds~\cite{olsen2018deepweeds}, 1.25 on PlantDoc~\cite{singh2019plantdoc}, and 14.53 on LUCAS~\cite{dandrimont2020lucas}. The corresponding exploratory New Plant Diseases difference is 1.06 points. Because only the 10\% regime and one seed were evaluated, these results describe label-limited transfer at one operating point rather than a label-efficiency curve.

CBRL has the largest macro-F1 point estimate on all four primary datasets, but the interpretation varies: the margins are large on DeepWeeds and LUCAS, only 0.43 points over Barlow Twins on near-saturated PlantVillage, and practically unhelpful on PlantDoc because all absolute scores are very low. These rankings should not be read as statistical superiority.

\subsection{Validation Dynamics}

\begin{figure}[t]
\centering
\begin{tikzpicture}
\begin{axis}[
width=0.98\columnwidth,
height=0.46\columnwidth,
xmin=0,
xmax=50,
ymin=0,
ymax=100,
xlabel={Fine-tuning epoch},
ylabel={Validation macro-F1 (\%)},
tick label style={font=\scriptsize},
label style={font=\scriptsize},
grid=major,
grid style={gray!25},
axis line style={gray!60},
legend columns=3,
legend style={font=\tiny,draw=none,at={(0.5,-0.20)},anchor=north,/tikz/every even column/.append style={column sep=0.16cm}}
]
\addplot[gray!70,very thick] coordinates {
  (1,10.49573869) (2,25.41424632) (3,33.14073086) (4,42.25370586) (5,47.13051915)
  (6,55.63636422) (7,60.48434973) (8,64.73188996) (9,67.30017066) (10,67.50018597)
  (11,70.92288733) (12,70.29899359) (13,73.23276997) (14,70.36334872) (15,75.07482171)
  (16,77.36480832) (17,77.61053443) (18,77.98212171) (19,81.60324693) (20,79.52539921)
  (21,81.09000325) (22,77.16992497) (23,81.07711673) (24,80.00627756) (25,83.27546120)
  (26,83.70510936) (27,82.96943903) (28,82.75830150) (29,83.27437639) (30,84.91662145)
  (31,84.54871178) (32,83.85735750) (33,84.59346890) (34,85.68555117) (35,85.83151102)
  (36,85.52803993) (37,85.74334383) (38,85.74655652) (39,86.22493148) (40,86.60125732)
  (41,87.02306151) (42,87.03470230) (43,87.27515340) (44,86.94732785) (45,87.19531894)
  (46,87.08156943) (47,86.90081239) (48,87.00776100) (49,86.99166179) (50,86.95775867)
};
\addlegendentry{Random-init ViT}
\addplot[methodcbrl,ultra thick] coordinates {
  (1,3.27460580) (2,4.40114513) (3,76.02053881) (4,86.60665154) (5,93.40471625)
  (6,94.60635185) (7,95.58781385) (8,96.85922861) (9,97.14035392) (10,96.93183899)
  (11,97.17414975) (12,96.93257809) (13,97.29889631) (14,97.69755006) (15,97.61617780)
  (16,97.84736037) (17,97.96368480) (18,97.77663350) (19,97.84173369) (20,98.07351232)
  (21,97.92668819) (22,97.85913229) (23,97.70863056) (24,98.20054770) (25,98.11058044)
  (26,97.96154499) (27,98.03068638) (28,97.78428674) (29,98.17583561) (30,98.16722870)
  (31,97.99464941) (32,98.13876748) (33,98.09611440) (34,98.06594849) (35,98.05942178)
  (36,98.10577035) (37,97.98980355) (38,98.02791476) (39,98.08058739) (40,98.06054831)
  (41,98.06399941) (42,98.12080860) (43,98.02134037) (44,98.02134037) (45,98.06280136)
  (46,98.06280136) (47,98.04716110) (48,98.04716110) (49,98.04716110) (50,98.04716110)
};
\addlegendentry{CBRL (ours)}
\addplot[methodagriarc,thick] coordinates {
  (1,0.48454842) (2,0.49429564) (3,2.35022511) (4,1.70779657) (5,11.65007502)
  (6,41.51648581) (7,65.35865664) (8,73.08686972) (9,74.99959469) (10,81.43070340)
  (11,81.65293932) (12,84.66837406) (13,84.32130814) (14,85.07292867) (15,85.02085209)
  (16,86.13907695) (17,86.58829927) (18,86.69090271) (19,87.08747625) (20,88.05709481)
  (21,88.15447688) (22,88.25120330) (23,88.54545951) (24,88.55872154) (25,88.92928958)
  (26,88.61470222) (27,89.64347839) (28,89.25330043) (29,89.63596821) (30,89.39837813)
  (31,89.80281353) (32,89.77519274) (33,89.70777988) (34,89.43482041) (35,89.62524533)
  (36,90.00475407) (37,90.03534913) (38,90.01562595) (39,89.91199136) (40,90.40443897)
  (41,90.34546018) (42,90.53683877) (43,90.57386518) (44,90.43123722) (45,90.32223225)
  (46,90.53435326) (47,90.42853117) (48,90.43682814) (49,90.43081999) (50,90.43081999)
};
\addlegendentry{ArcFace-SSL}
\addplot[methodhific,thick] coordinates {
  (1,1.63076110) (2,8.90347511) (3,22.23696709) (4,36.50297821) (5,44.38178241)
  (6,57.32880235) (7,67.11206436) (8,69.78332996) (9,71.60471678) (10,78.36728692)
  (11,79.21330929) (12,81.76264763) (13,83.49518180) (14,83.75575542) (15,83.31828713)
  (16,84.93741155) (17,85.40419340) (18,86.89492941) (19,87.56935000) (20,88.55748177)
  (21,87.10334897) (22,88.64019513) (23,88.39729428) (24,88.69329691) (25,89.63467479)
  (26,88.78855109) (27,89.22778964) (28,88.85614872) (29,89.73304033) (30,90.04693627)
  (31,90.59266448) (32,89.05625939) (33,90.33067226) (34,90.03956914) (35,90.20389915)
  (36,90.84268212) (37,90.53142071) (38,90.32532573) (39,90.90256095) (40,90.97589850)
  (41,91.51796103) (42,91.58788919) (43,91.24863148) (44,91.07989669) (45,90.99540114)
  (46,91.15588069) (47,91.11736417) (48,91.15221500) (49,91.13573432) (50,91.13838077)
};
\addlegendentry{HiFiC-SSL}
\addplot[gray!55,dashed,thick] coordinates {
  (1,3.25849466) (2,5.46031483) (3,8.88508558) (4,12.42451519) (5,15.67667872)
  (6,18.54558140) (7,20.68526298) (8,21.44740522) (9,23.17832112) (10,23.57395589)
  (11,24.09242392) (12,25.26074052) (13,26.08874142) (14,25.15826821) (15,25.98962784)
  (16,26.73540413) (17,26.94194317) (18,27.24668384) (19,27.80061662) (20,28.21806967)
  (21,28.30494046) (22,28.54155302) (23,29.24565375) (24,28.72161269) (25,29.26043570)
  (26,29.59179878) (27,29.77611125) (28,29.94123101) (29,29.63587344) (30,30.29836714)
  (31,30.58207929) (32,30.89280128) (33,30.42063713) (34,30.64769506) (35,30.37728369)
  (36,31.04438186) (37,30.76544106) (38,31.04939461) (39,31.38590157) (40,31.10720515)
  (41,31.24973774) (42,31.35351837) (43,31.36702776) (44,31.22844994) (45,31.61598444)
  (46,31.24634027) (47,31.32795095) (48,31.30439222) (49,31.31616116) (50,31.31616116)
};
\addlegendentry{DINOv2}
\addplot[methodbyol,dashdotted,thick] coordinates {
  (1,0.73108999) (2,2.41869297) (3,10.51045060) (4,47.81423807) (5,68.32338572)
  (6,77.95885205) (7,80.96628785) (8,83.05108547) (9,85.61322093) (10,86.41468883)
  (11,87.92563677) (12,88.48235011) (13,88.73267174) (14,89.15539980) (15,89.97756839)
  (16,90.17173052) (17,90.38224816) (18,90.61785340) (19,90.37944674) (20,90.93569517)
  (21,90.73185921) (22,91.21849537) (23,91.29761457) (24,91.56525731) (25,92.00687408)
  (26,92.16502309) (27,92.04960465) (28,91.93189144) (29,92.28060842) (30,92.24115610)
  (31,92.23227501) (32,92.57120490) (33,92.29145646) (34,92.56089926) (35,92.36307144)
  (36,92.57259965) (37,92.56210327) (38,92.33562350) (39,92.52126813) (40,92.49803424)
  (41,92.47917533) (42,92.66079664) (43,92.77935028) (44,92.56919622) (45,92.65592098)
  (46,92.70973802) (47,92.68886447) (48,92.70783663) (49,92.74300933) (50,92.72789955)
};
\addlegendentry{BYOL}
\addplot[methoddenoising,thick] coordinates {
  (1,3.46394889) (2,20.69388330) (3,36.78363562) (4,52.26563811) (5,61.64909601)
  (6,68.89700294) (7,72.59553671) (8,72.59628773) (9,80.12326360) (10,83.71324539)
  (11,83.70734453) (12,84.45426822) (13,84.48457122) (14,85.07837057) (15,87.06015348)
  (16,88.90839219) (17,86.56799793) (18,87.58000731) (19,88.99102807) (20,89.56052065)
  (21,88.10473680) (22,88.85691166) (23,88.91365528) (24,89.99850154) (25,91.06622338)
  (26,90.47841430) (27,90.96354842) (28,90.51996469) (29,91.32568836) (30,90.89475870)
  (31,91.95237160) (32,90.19389153) (33,91.06751680) (34,91.49265885) (35,91.89876914)
  (36,92.00060964) (37,92.05611944) (38,91.88606143) (39,91.73792601) (40,92.54199266)
  (41,92.09709167) (42,92.42190719) (43,92.14992523) (44,92.28622317) (45,92.46537089)
  (46,92.48666167) (47,92.42154956) (48,92.28317738) (49,92.27295518) (50,92.27295518)
};
\addlegendentry{Denoising}
\addplot[methodmae,thick] coordinates {
  (1,0.48887315) (2,3.16060148) (3,6.19506426) (4,29.06932533) (5,47.92598784)
  (6,59.45533514) (7,70.03016472) (8,76.39971972) (9,78.51949334) (10,81.68489933)
  (11,84.15873647) (12,85.67407727) (13,86.14485860) (14,88.30993176) (15,89.33211565)
  (16,88.92683983) (17,90.16286135) (18,90.60082436) (19,89.89747167) (20,90.96587300)
  (21,90.96574187) (22,90.79093337) (23,91.55600667) (24,91.76667333) (25,91.82167649)
  (26,92.55761504) (27,92.60511398) (28,92.31070280) (29,92.77650118) (30,92.64678359)
  (31,92.83936620) (32,92.78556108) (33,92.96973348) (34,93.17176938) (35,92.35146046)
  (36,92.84612536) (37,92.75367856) (38,92.89451838) (39,93.16820502) (40,92.95628071)
  (41,93.00912023) (42,93.22742820) (43,93.03967357) (44,93.15134287) (45,93.26413870)
  (46,93.15384626) (47,93.13734174) (48,93.11826825) (49,93.11826825) (50,93.11826825)
};
\addlegendentry{MAE}
\addplot[methodsimclr,thick] coordinates {
  (1,16.43117517) (2,51.90598965) (3,75.19650459) (4,85.58142781) (5,88.73808384)
  (6,91.08508825) (7,92.03115106) (8,92.70134568) (9,93.18286777) (10,93.39065552)
  (11,94.56373453) (12,93.48617792) (13,93.52407455) (14,93.62648726) (15,94.50846314)
  (16,94.31056976) (17,94.48359013) (18,94.89603043) (19,93.87078881) (20,94.73965168)
  (21,95.10227442) (22,95.10264397) (23,95.12726665) (24,94.64324117) (25,95.23929358)
  (26,94.96861100) (27,94.50117350) (28,95.04700899) (29,95.32250166) (30,95.29553652)
  (31,95.45799494) (32,95.33280730) (33,94.71093416) (34,95.55547833) (35,95.64197063)
  (36,95.66856623) (37,95.47250867) (38,95.43926120) (39,95.80819607) (40,95.68161964)
  (41,95.95450163) (42,95.89874148) (43,96.02248669) (44,95.92880607) (45,95.89294791)
  (46,95.96225023) (47,96.01154327) (48,95.94567418) (49,95.93146443) (50,95.93146443)
};
\addlegendentry{SimCLR}
\addplot[black,densely dotted,thick] coordinates {
  (1,3.61670032) (2,4.82394658) (3,48.35331440) (4,84.17428732) (5,92.50734448)
  (6,93.73700023) (7,94.39642429) (8,94.94752288) (9,95.69330812) (10,95.85304856)
  (11,95.76175213) (12,95.83896995) (13,96.13503218) (14,96.51451111) (15,96.10109329)
  (16,96.66997194) (17,96.57984972) (18,96.57794833) (19,96.69157863) (20,96.75567150)
  (21,96.67691588) (22,97.30346799) (23,96.60019279) (24,96.99543715) (25,97.19836712)
  (26,96.75241709) (27,96.99761271) (28,96.52099013) (29,96.88188434) (30,96.98126912)
  (31,97.13642001) (32,96.76126242) (33,97.09643126) (34,97.00357318) (35,97.16765881)
  (36,97.18022943) (37,97.14974761) (38,96.96218371) (39,96.94635272) (40,97.13909030)
  (41,97.08598256) (42,97.13259339) (43,97.22018242) (44,97.22685218) (45,97.27294445)
  (46,97.19817638) (47,97.21200466) (48,97.31879830) (49,97.31293917) (50,97.31293917)
};
\addlegendentry{Barlow Twins}
\end{axis}
\end{tikzpicture}
\caption{Descriptive validation macro-F1 trajectories for all ten single-run PlantVillage experiments under the common 10\%-label fine-tuning protocol.}
\label{fig:validation-curves}
\end{figure}
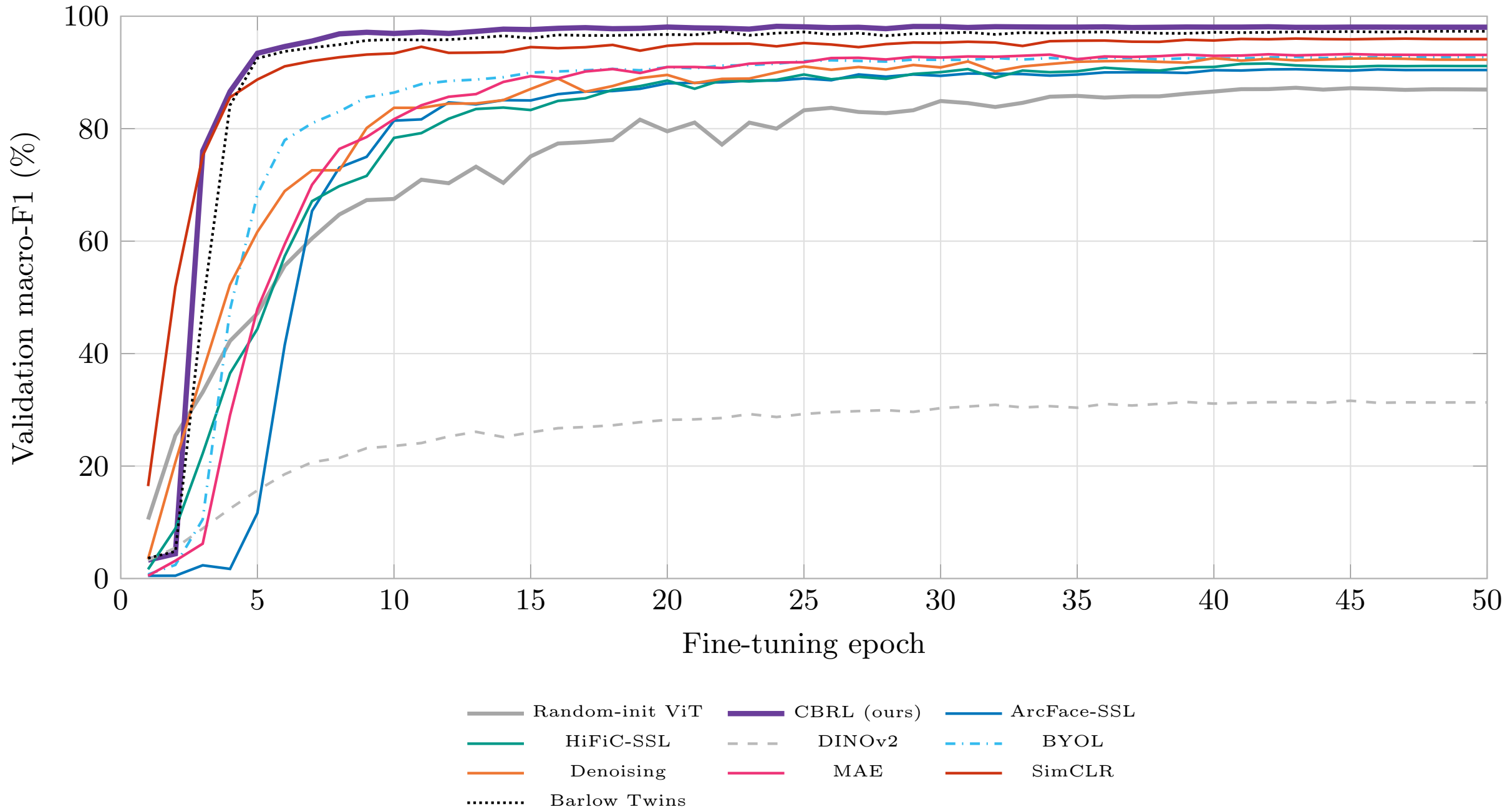

Figure~\ref{fig:validation-curves} shows the PlantVillage~\cite{hughes2015plantvillage} validation macro-F1 trajectories during downstream fine-tuning with 10\% of the training labels. These curves provide a descriptive view of optimisation behaviour and checkpoint selection under the common 50-epoch fine-tuning protocol.

In its single run, CBRL reaches the selected validation macro-F1 of 98.20\% at epoch 24. Near that checkpoint, its trajectory lies above the other nine displayed PlantVillage~\cite{hughes2015plantvillage} runs. This is a descriptive optimisation trace, not evidence of convergence stability across runs.

\section{Discussion}

CBRL records the largest macro-F1 point estimate on each of the four primary benchmarks. This pattern motivates further evaluation of the complete formulation, but a single run cannot establish statistical superiority, robustness, or synergy between its objectives. The 33-image New Plant Diseases result is exploratory and does not strengthen the primary cross-dataset claim.

The parent methods provide useful but limited context. Within each dataset-specific comparison, all three used ViT-L/16, the same unlabelled images, 10,000 steps, batch size 256, and the same downstream protocol. BYOL additionally matches CBRL's online/target architecture and most pretraining settings; SimCLR uses its method-native projector, optimiser, and augmentations. CBRL's higher recorded values therefore show how the implemented formulations compared, not that adding the two losses caused the difference.

The random-initialisation, label-only condition asks whether SSL initialisation helps relative to training ViT-L/16 from scratch with scarce labels. Its weakness is understandable for such a large model and does not make it a comprehensive supervised-transfer reference. In particular, the study does not compare against ImageNet-supervised initialisation or a separately tuned supervised recipe.

PlantDoc is an important negative result: all methods remain at failure-level performance, including CBRL at 9.31\% macro-F1. On LUCAS, the accuracy--macro-F1 gap indicates that aggregate and class-balanced performance differ substantially, but aggregate metrics alone do not support a claim about minority classes; class-wise support and error analysis would be required.

DINOv2 and Barlow Twins also require cautious interpretation. Their unusually low values in some datasets may reflect implementation, optimisation, augmentation, or instability, but the available logs and downstream scores do not isolate a cause. These runs should neither be silently discarded nor used to conclude that the general methods are unsuitable for agricultural imagery.

\section{Limitations}

Several limitations constrain the interpretation of the results.

First, the experiments use one random seed and one final test evaluation per method on the five test datasets. The comparison is a controlled point-estimate benchmark, but it does not quantify variation caused by initialisation, data sampling, or stochastic optimisation.

Second, statistical significance is not reported. Repeating each regime across multiple random seeds would enable means, standard deviations, confidence intervals, and paired comparisons.

Third, the same fine-tuning hyperparameters are used across datasets with very different sizes and class distributions. This controls the comparison but may disadvantage methods or datasets that require different regularisation, augmentation, or learning-rate schedules.

Fourth, pretraining curves are intentionally excluded because the quantitative scope is limited to downstream classification fine-tuning. The unusually low Barlow Twins results on DeepWeeds and PlantDoc motivate a separate analysis of pretraining stability and representation collapse.

Fifth, the New Plant Diseases held-out test split contains only 33 images and represents only a subset of the 38 output classes. Because the implemented macro-F1 averages across all 38 classes, classes absent from this test split contribute zero; consequently, even perfect predictions yield 21.05\% macro-F1. These test results should therefore be interpreted as limited evidence from a small, incompletely representative split.

Sixth, the experiment contains no strict CBRL loss-toggle ablation or matched weighting study. BYOL is closely matched to CBRL, whereas SimCLR retains method-native pretraining settings. The results establish the point estimates of the implemented formulations, not the individual contribution or synergy of NT-Xent and EMA-target prediction. The 1.0/1.0 coefficients are a fixed, non-tuned choice and do not establish optimal balance.

Seventh, the random-initialisation baseline isolates the value of SSL relative to learning ViT-L/16 from scratch with limited labels, but it is not a substitute for supervised-transfer baselines such as ImageNet initialisation.

\section{Conclusion}

This study evaluates CBRL, a specific joint contrastive and EMA-target bootstrap formulation, for 10\%-label agricultural classification. In single-seed runs, CBRL records the largest macro-F1 point estimate on four primary benchmarks, with its largest margin over random initialisation on DeepWeeds. The evidence is deliberately bounded: PlantDoc is a negative result at failure-level absolute performance, New Plant Diseases is only exploratory, the baseline does not cover supervised transfer, and no strict loss-toggle or repeated-seed analysis is available. CBRL should therefore be viewed as a reproducible candidate formulation that warrants confirmatory evaluation, not as evidence of objective synergy, statistical superiority, or established robustness.

\backmatter

\bmhead{Data availability}

The five datasets considered in this study are publicly available from their original providers, as cited in the dataset descriptions. The aggregate held-out test results needed to support the findings are reported in this paper.

\bmhead{Acknowledgements}

The authors gratefully acknowledge Precision AI for providing the GPU computing resources used to train the models and conduct the experiments.

\bibliography{references}

\end{document}